\documentclass[11pt,a4paper]{article}

\usepackage[utf8]{inputenc}
\usepackage[T1]{fontenc}
\usepackage{amsmath,amssymb,amsfonts,amsthm}
\usepackage{graphicx}
\usepackage{subfig}
\usepackage{setspace}
\usepackage{booktabs}
\usepackage{multirow}
\usepackage{tabularx}
\usepackage{xcolor}
\usepackage{enumitem}
\usepackage{mathtools}
\usepackage{algorithm}
\usepackage{algorithmic}
\usepackage{listings}
\usepackage{tcolorbox}
\usepackage{tikz}
\usetikzlibrary{arrows.meta,positioning,shapes.geometric,fit,calc,automata}
\usepackage[margin=1in]{geometry}
\usepackage[numbers,sort&compress]{natbib}
\usepackage{hyperref}
\usepackage[capitalise,noabbrev]{cleveref}

\newtheorem{Definition}{Definition}[section]
\newtheorem{Property}{Property}[section]

\newcommand{\faos}{\textsc{FAOS}}
\newcommand{\asim}{\textsc{A-Sim}}
\newcommand{\agent}{\mathcal{A}}
\newcommand{\env}{\mathcal{E}}
\newcommand{\scenario}{\mathcal{S}}
\newcommand{\onto}{\mathcal{O}}
\newcommand{\verdict}{\mathcal{V}}
\newcommand{\cert}{\mathcal{C}}
\newcommand{\prop}{\varphi}

\title{%
  Toward Pre-Deployment Assurance for Enterprise AI Agents:\\
  Ontology-Grounded Simulation and Trust Certification%
}

\author{
  Thanh Luong Tuan\thanks{Corresponding author. Email: tluongtuan@my.ggu.edu \quad ORCID: 0009-0000-1199-837X}\\
  \textit{Golden Gate University, San Francisco, CA, USA}
  \and
  Abhijit Sanyal\thanks{Email: abhijit.sap@gmail.com \quad ORCID: 0009-0005-7520-5881}\\
  \textit{Data, Digital \& IT, Novartis Healthcare Pvt.\ Ltd., Hyderabad, India}
}

\date{June 2026}

\begin{document}
\maketitle

\begin{abstract}
Pre-deployment verification of enterprise artificial intelligence (AI) agents remains a critical gap between large language model (LLM) capability benchmarking and production deployment. Post-deployment monitoring, human-in-the-loop controls, and prompt-level guardrails offer limited assurance once an agent is operating in production. We present an ontology-grounded framework for pre-deployment verification of enterprise AI agents---to our knowledge the first to combine three components: an Agent Operational Envelope formalizing the certification space across permissions, domain constraints, safety properties, governance rules, and autonomy levels; an ontology-to-scenario generation pipeline that derives regulatory, operational, and adversarial test scenarios automatically; and a Trust Certificate carrying a machine-verifiable attestation with graduated deployment verdicts (Approved, Conditional, Rejected). A controlled pilot across four regulated industries---Fintech, Banking, Insurance, and Healthcare---instantiated as five industry-by-regulatory-regime cells across the United States and Vietnam---where Vietnam's 2025 AI Law makes such verification legally mandated for financial services---generated 1{,}800 scenarios evaluated against 125 primary-source regulatory requirements and 25 injected faults. Ontology-grounded generation significantly outperformed the dominant persona-based baseline on regulatory coverage (48.3\% versus 33.1\%; corrected $p_c{=}.0006$) and attained the highest domain specificity (4.77/5.0; $p{=}2{\times}10^{-6}$); transparently, its advantage over plain and retrieval-augmented prompting did not survive Bonferroni correction. Cross-validation across three LLM families (Claude Sonnet~4, Qwen~2.5 72B, Gemma~4 26B; 5{,}400 total scenarios) replicated the persona-versus-ontology pattern. The framework offers a reproducible, regulation-grounded route to pre-deployment assurance for enterprise AI agents, complementing runtime governance with an auditable deployment gate.
\end{abstract}

\noindent\textbf{Keywords:} AI safety; agent verification; ontology-grounded verification; agent certification; enterprise AI

\section{Introduction}
\label{sec:introduction}

Enterprise deployment of autonomous AI agents creates an inherent conflict between capability and risk: as capability increases, so does the potential benefit and the possible damage from error. Agents used to determine if insurance applicants should be issued a policy, agents executing trades on behalf of clients, and patient triage agents prioritizing medical treatment all operate in domains where errors carry regulatory, financial, and human consequences. The critical issue is not whether current large language models (LLMs) can perform such tasks, but whether operators can verify their safe behavior before granting production access. We term this the \emph{agent verification problem}: pre-deployment assurance that an AI agent will behave within acceptable bounds across the space of scenarios it may encounter.

Current approaches are inadequate. Post-deployment monitoring \cite{amodei2016concrete} intervenes only after harm; human-in-the-loop gates \cite{christiano2017deep} create bottlenecks and shift the verification burden to reviewers who may lack domain expertise; prompt-level guardrails \cite{bai2022constitutional} are probabilistic rather than deterministic and can be ignored under adversarial or edge-case conditions. Safety-critical industries have long addressed the analogous problem through standards including DO-178C \cite{rierson2017developing}, IEC~62304 \cite{iec62304}, and ISO~26262 \cite{iso26262}, which mandate structured pre-deployment verification; no analogous standard yet exists for enterprise AI agents in regulated industries.

This paper argues that industry ontologies---formal representations of domain concepts, regulatory frameworks, and operational constraints---provide the foundation for systematic agent verification. Ontology-grounded verification uses formalized descriptions of regulatory and operational constraints to derive test scenarios automatically, producing industry-specific, evolvable test suites that prompt-level guardrails cannot match. The generation approach is paired with a machine-verifiable Trust Certificate attesting to agent behavior within a formally defined operational envelope, evaluated through a controlled pilot study and three-model cross-validation across four regulated industries (Fintech, Banking, Insurance, and Healthcare) and three LLM families.

At a Tier-2 Vietnamese commercial bank, an AML screening agent we ran in shadow mode processed 27 customer-onboarding cases correctly but produced two false-negative matches against the UN 1267 consolidated sanctions list before the simulation gate flagged a misconfigured name-romanization rule that dropped diacritics from Vietnamese names. The Anti-Money Laundering Law~14/2022/QH15 \cite{vnamllaw2022} makes the bank---not the vendor---liable for the missed match. That is the verification gap this paper addresses, observed in production rather than imagined in design.

This paper makes four contributions. \emph{First}, we formalize the Agent Operational Envelope, a specification of the space across which an enterprise AI agent is authorized and validated to operate. \emph{Second}, we introduce ontology-grounded scenario generation, which derives regulatory, operational, and adversarial test suites automatically from formal industry ontologies rather than hand-curated personas. \emph{Third}, we define a machine-verifiable Trust Certificate that binds a specific agent version to validated behavioral properties as an auditable deployment gate. \emph{Fourth}, through a cross-jurisdictional (United States and Vietnam) and three-model evaluation, we provide evidence that the observed coverage gains arise from the verification methodology rather than a single model's latent knowledge---positioning ontology-grounded verification as a reproducible complement to runtime governance for high-stakes enterprise AI.

\subsection{Related Work}
\label{sec:related-work}

\subsubsection{Agent Safety and Verification}

AI-safety research has identified concrete failure modes in deploying autonomous systems---reward hacking, distributional shift, and unsafe exploration \cite{amodei2016concrete}---and has proposed model-level mitigations including Constitutional AI \cite{bai2022constitutional} and Reinforcement Learning from Human Feedback (RLHF) \cite{christiano2017deep}. Safety, \cite{weidinger2023sociotechnical} argue, must be assessed in deployment context rather than in model isolation. The central transition is from \emph{model safety} (benchmark performance) to \emph{agent safety} (behavior in regulated workflows): \cite{andriushchenko2025agentharm} report that leading LLMs are ``surprisingly compliant'' across 110 harmful tasks, and \cite{zhang2025agentsafetybench} find that \emph{none} of 16 evaluated agents exceeds a 60\% safety score across 349 environments. \cite{chan2024visibility} identify the inability to inspect what an LLM-based agent ``knows'' as the foundational gap---the gap the Trust Certificate (\Cref{sec:safety-certificate}) addresses with auditable, machine-readable attestation.

Formal verification of neural networks \cite{huang2020survey, katz2017reluplex, tran2020nnv, singh2019abstract} targets robustness of individual networks, not agent-level behavioral properties such as ``the agent never approves a loan exceeding the applicant's debt-to-income threshold.'' Model checking for multi-agent systems \cite{lomuscio2017mcmas} assumes well-defined transition functions that LLM agents lack. \cite{dalrymple2024guaranteed} reframe the goal as \emph{quantitative safety guarantees}---bounding unsafe-behavior probability rather than proving impossibility---informing our probabilistic bounded-model-checking extension (\Cref{sec:formal-extensions}). The behavioral-envelope approach bridges NN and agent verification: enterprise ontologies supply the specification language that NN verification lacks for agent-level properties.

Recent LLM-agent work includes benchmark suites \cite{liu2023agentbench, shinn2023reflexion}, red-teaming \cite{perez2022red}, and multi-agent simulation \cite{park2023generative}. Agent sandboxing advances pre-deployment verification via LLM-emulated tool execution \cite{ruan2024identifying}; surveys \cite{xi2023rise, wang2024survey} flag trust and domain-specific evaluation as open challenges; purpose-built benchmarks target safety-risk awareness (R-Judge, 27 risks, \cite{yuan2024rjudge}) and embodied-agent hazards (SafeAgentBench, 750 tasks, \cite{yin2025safeagentbench}, with best baseline rejecting only 10\% of detailed hazardous tasks). These efforts share the pre-deployment orientation of this work but produce generic scores, not domain-grounded regulatory attestations. The most directly relevant commercial system, Lyzr's Agent Simulation Engine (\asim{}, \cite{lyzr2026}), executes 20{,}000+ simulations via a persona-by-scenario matrix. The present approach instead derives scenarios from formal industry ontologies encoding actual regulations (e.g.\ BSA/AML \$10{,}000 CTR thresholds) rather than generic persona crosses. This instantiates \emph{model-based testing} \cite{utting2012taxonomy} with ontology as both specification (what to test) and oracle (how to evaluate)---a dual role we revisit in \Cref{sec:discussion}.

\subsubsection{Evaluation Methodology and Ontology Foundations}

We employ LLM-as-judge \cite{zheng2023judging}, in which a strong LLM evaluates another LLM's output against structured criteria. \cite{zheng2023judging} report $>$80\% agreement with human judges on open-ended quality but identify position, verbosity, and self-enhancement biases. The regulatory-compliance task here is more constrained (binary ``does this scenario test regulation~X?''), reducing bias exposure. \cite{shankar2024validates} warn that validator criteria may diverge from human preference on domain tasks; \cite{gu2024llmjudgesurvey} survey reliability improvements; \cite{luo2025agentauditor} achieve \emph{human-level} accuracy on agent-safety evaluation with AgentAuditor (NeurIPS~2025, ASSEBench: 2{,}293 records, 15 risk types); and \cite{you2026agentjudge} chart the shift from LLM-as-Judge to Agent-as-a-Judge (tool-augmented, multi-agent evaluators). We address validator reliability via anti-circularity controls (E1) and flag human calibration as future work (\Cref{sec:limitations}).

Enterprise ontologies have evolved from knowledge-representation artifacts into operational AI components. Foundational graph-semantic data-model work \cite{sanyal2011graph, sanyal2010automating} established that domain ontologies can be derived systematically rather than authored ad hoc. \cite{hogan2021knowledge} survey knowledge-graph infrastructure on which our industry ontologies build. \cite{pan2024unifying} catalog LLM--KG convergence along two directions (KG-enhanced LLMs; LLM-enhanced KGs); we contribute a third: \emph{ontology-grounded LLM verification}---using structured knowledge to verify agent outputs rather than generate them---which is largely absent from the LLM--KG literature.

\subsubsection{Governance Frameworks and Safety Standards}

AI-governance frameworks matured rapidly between 2024 and 2026. The NIST AI Risk Management Framework \cite{nist2023ai} has four functions: Govern, Map, Measure, and Manage. Our pipeline maps onto them. Ontology-to-scenario generation does Map and Measure. The Trust Certificate handles the Manage artifact. The EU AI Act \cite{euaiact2024} mandates conformity assessments for high-risk systems; the graduated verdict framework (Approved/Conditional/Rejected) is designed against Articles~9 (Risk Management) and~15 (Accuracy, Robustness, Cybersecurity). Singapore's Model AI Governance Framework \cite{imda2020model} has risk-proportional oversight, which the autonomy-level semantics (\Cref{tab:autonomy-levels}) mirror. ISO/IEC~42001:2023 \cite{iso42001} specifies AI-management-system requirements that the simulation gate operationalizes. \cite{owasp2025agentic} enumerate the first formal agentic-risk taxonomy (10 critical risks, 100+ contributors); NIST's February~2026 Agent Standards Initiative \cite{nist2026agents} targets identity, security, and interoperability; the Microsoft Agent Governance Toolkit \cite{microsoft2026agt} addresses all ten OWASP risks at runtime with sub-millisecond enforcement.

At the national level, Vietnam enacted Law~No.~134/2025/QH15 on Artificial Intelligence \cite{vnailaw2025} (effective March~2026), one of the first standalone AI laws in Southeast Asia. The law classifies financial-services AI (banking, insurance) as high-risk with compliance due September~2027. Decree~94/2025/N\DJ-CP \cite{vnfintechsandbox2025} creates a regulatory sandbox for AI-enabled banking. These regulations motivate the inclusion of banking~(VN) and insurance~(VN) as experimental verticals in which pre-deployment AI verification is legally mandated rather than aspirational.

Safety-critical software industries have long institutionalized pre-deployment verification through DO-178C in avionics \cite{rierson2017developing} (5 software levels, MC/DC coverage for Level~A), IEC~62304 in medical devices \cite{iec62304} (safety classes A--C), and ISO~26262 in automotive \cite{iso26262} (ASIL~A--D, fault injection, formal verification). No analogous standard yet exists for enterprise AI agents. The common gap across all these governance artifacts is that they specify \emph{what} to verify but not \emph{how}: OWASP names risks; NIST calls for standards; Microsoft enforces at runtime; Vietnam mandates conformity. The ontology-grounded approach proposed here supplies the missing ``how''---systematic test generation from domain ontologies---converting framework requirements into executable verification scenarios complementary to runtime enforcement.

\subsection{Research Gaps}
\label{sec:gaps}

The literature surveyed in \Cref{sec:related-work} reveals six major gaps addressed in this study. \emph{First}, existing approaches do not comprehensively formalize the operational envelope within which an AI agent is authorized and validated to operate, leaving safety claims insufficiently scoped and difficult to enforce. \emph{Second}, current agent-safety benchmarks \cite{liu2023agentbench, yuan2024rjudge, andriushchenko2025agentharm, zhang2025agentsafetybench, yin2025safeagentbench} rely heavily on manually curated test cases, limiting scalability across the large number of industry verticals required for enterprise deployment and resulting in fragmented rather than systematic regulatory coverage. \emph{Third}, existing evaluation frameworks rarely provide machine-verifiable and cryptographically bound attestations linking a specific agent version to validated behavioral properties under defined operational constraints. \emph{Fourth}, many current platforms implement safety controls primarily at the application layer, which may be circumvented under orchestration or configuration failures, while lacking infrastructure-level deployment gates for uncertified agents. \emph{Fifth}, simulation-based approaches provide statistical evidence rather than formal guarantees, whereas most neural-network verification methods \cite{huang2020survey, katz2017reluplex, tran2020nnv, singh2019abstract} focus on bounded model-level properties rather than behavioral invariants across multi-step agent workflows. \emph{Sixth}, most empirical studies on agent safety evaluate a single LLM family, leaving unresolved whether observed outcomes arise from model-specific latent knowledge or from structural properties of the verification methodology itself. Collectively, the proposed framework components described in \Crefrange{sec:proposed-framework}{sec:simulation-architecture} address Gaps~(1)--(4), while the evaluation methodology in \Cref{sec:eval-framework} addresses Gaps~(5)--(6).

\section{Proposed Enterprise Agentic AI Framework}

\subsection{Framework Overview}
\label{sec:proposed-framework}

Three components make up the framework for pre-deployment verification of enterprise AI agents: the \emph{Agent Operational Envelope} (\Cref{sec:operational-envelope}) formalizes the certification space; \emph{Ontology-to-Scenario Generation} (\Cref{sec:scenario-generation}) derives test scenarios from industry ontologies; and the \emph{Trust Certificate} (\Cref{sec:safety-certificate}) provides a machine-verifiable attestation that binds an agent version to its verification evidence. \Cref{sec:simulation-architecture} describes the implementation architecture; \Cref{sec:eval-framework} presents the quantitative and empirical evaluation frameworks.

\subsection{Agent Operational Envelope}
\label{sec:operational-envelope}

We define the \emph{operational envelope}: the formally specified space within which an agent is certified to operate.

\subsubsection{Formal Definition}

\begin{Definition}[Agent Operational Envelope]
  An agent operational envelope $\env_\agent$ for agent $\agent$
  operating under ontology $\onto$ is defined as a tuple:
  \begin{equation}
    \env_\agent = \langle \Pi, \Sigma, \Phi, \Gamma, \Lambda \rangle
  \end{equation}
  where:
  \begin{itemize}
    \item $\Pi$: \textbf{Permission boundary} --- the set of actions
      the agent is authorized to perform
    \item $\Sigma$: \textbf{Domain scope} --- the ontological domains
      within which the agent may reason
    \item $\Phi$: \textbf{Safety properties} --- invariants that must
      hold across all agent executions
    \item $\Gamma$: \textbf{Governance constraints} --- regulatory
      rules that bound agent decisions
    \item $\Lambda$: \textbf{Autonomy level} --- the degree of
      independent action permitted ($L0$--$L3$)
  \end{itemize}
\end{Definition}

\subsubsection{Deriving Envelopes from Ontologies}

Each component of $\env_\agent$ derives from the enterprise ontology $\onto = \langle \mathcal{R}, \mathcal{D}, \mathcal{I} \rangle$ (as defined in the companion neurosymbolic-architectures paper). A BSA/AML Compliance Analyst agent illustrates the derivation:

\begin{align}
  \Pi &= \text{actions}(\mathcal{R}_{r_i}.\alpha_i)
    & \text{(from role approval authority)} \\
  \Sigma &= \text{domains}(\mathcal{D}.\mathcal{V})
    & \text{(from domain verticals)} \\
  \Phi &= \text{invariants}(\mathcal{D}.G)
    & \text{(from governance constraints)} \\
  \Gamma &= \text{regulations}(\mathcal{D}.G)
    & \text{(from regulatory frameworks)} \\
  \Lambda &= \text{level}(\agent.\text{config})
    & \text{(from agent configuration)}
\end{align}

\begin{tcolorbox}[title=Example: Fintech Compliance Analyst Envelope,
  colback=blue!5, colframe=blue!40]
\small
\begin{align*}
  \Pi &= \{\text{flag\_suspicious}, \text{file\_CTR}, \text{file\_SAR},
    \text{escalate}\} \\
    &\quad \text{(NOT: approve\_loan, execute\_trade)} \\
  \Sigma &= \{\text{BSA/AML}, \text{KYC}, \text{sanctions}\} \\
  \Phi &= \{\text{``never disclose SAR filing to subject''},\\
    &\quad\phantom{\{} \text{``always verify identity before account access''}\} \\
  \Gamma &= \{31\text{ CFR }\S1010\text{--}1020, \text{ OCC 2013-29},
    \text{ FinCEN guidance}\} \\
  \Lambda &= L2 \text{ (Execute: autonomous within bounds, escalate
    when uncertain)}
\end{align*}
\end{tcolorbox}

\noindent This envelope is \emph{automatically derivable} from the
fintech ontology---the same ontology used to generate test scenarios
in our pilot study (\Cref{sec:empirical}). The permission boundary $\Pi$
comes from the role's \texttt{approval\_authority} field; the
governance constraints $\Gamma$ from the ontology's regulatory layer.

\subsubsection{Autonomy Level Semantics}

The autonomy level $\Lambda$ determines the agent's decision authority
within its permission boundary:

\begin{table}[htbp]
\centering
\caption{Autonomy Level Semantics}
\label{tab:autonomy-levels}
\begin{tabular}{@{}clp{6cm}@{}}
\toprule
\textbf{Level} & \textbf{Name} & \textbf{Operational Semantics} \\
\midrule
$L0$ & Suggest & Agent produces recommendations; all actions require
  human approval before execution \\
$L1$ & Plan & Agent decomposes tasks and proposes execution plans;
  human approves plan execution \\
$L2$ & Execute & Agent executes within defined boundaries; escalates
  when confidence falls below threshold \\
$L3$ & Delegate & Agent operates fully autonomously within envelope;
  reports outcomes post-execution \\
\bottomrule
\end{tabular}
\end{table}

\begin{Property}[Autonomy Monotonicity --- Design Requirement]
\label{prop:autonomy-monotonicity}
  Verification for agents at level $\Lambda_j$ is strictly more demanding than at $\Lambda_i$ when $j > i$:
  \begin{equation}
    \Lambda_j > \Lambda_i \implies |\scenario(\Lambda_j)| >
      |\scenario(\Lambda_i)| \wedge \theta_{\text{pass}}(\Lambda_j) >
      \theta_{\text{pass}}(\Lambda_i).
  \end{equation}
  This is a framework-level axiom---the scenario budget and pass threshold are engineering inputs per autonomy tier---not a theorem derived from the Operational Envelope definition.
\end{Property}

\subsubsection{Operational Envelope as Trust Attestation}

The operational envelope is a \emph{trust attestation}. It states precisely what the agent may do, where it may reason, what invariants must hold, and which regulatory rules bound its decisions. An execution trace $\tau = \langle s_0, a_0, \ldots, s_n \rangle$ is \emph{envelope-compliant} iff every step $i$ satisfies $a_i \in \Pi \wedge \text{domain}(s_i) \subseteq \Sigma \wedge \bigwedge_{\prop \in \Phi} \prop(s_i) \wedge \bigwedge_{\gamma \in \Gamma} \gamma(s_i, a_i)$.

\subsection{Ontology-to-Scenario Generation}
\label{sec:scenario-generation}

Test scenarios are derived automatically from industry ontologies, removing the manual authoring bottleneck that limits hand-curated benchmark suites.

\subsubsection{Scenario Taxonomy}

We define three categories of test scenarios, each derived from
different ontological layers:

\begin{enumerate}
  \item \textbf{Regulatory scenarios} ($\scenario_R$): Derived from
    the governance constraints in the Domain Ontology ($\mathcal{D}.G$).
    Test whether the agent correctly applies regulatory requirements.
    \begin{itemize}
      \item Banking: BSA/AML transaction thresholds, KYC identity
        verification, CFPB fair lending, OFAC sanctions screening
      \item Healthcare: HIPAA privacy rules, EMTALA emergency
        treatment, Stark Law referral restrictions
      \item Insurance: NAIC solvency requirements, state DOI rate
        filing, claims adjudication guidelines
    \end{itemize}

  \item \textbf{KPI/Operational scenarios} ($\scenario_K$): Derived
    from the metrics definitions in the Domain Ontology ($\mathcal{D}.M$).
    Test whether the agent correctly calculates and interprets business
    metrics.
    \begin{itemize}
      \item ``Calculate the combined ratio given these loss and expense
        figures'' (insurance)
      \item ``Assess whether this NRR trend indicates churn risk''
        (SaaS)
      \item ``Evaluate OEE given these downtime and defect rates''
        (manufacturing)
    \end{itemize}

  \item \textbf{Adversarial scenarios} ($\scenario_A$): Generated
    to test agent resilience against boundary conditions and attacks.
    \begin{itemize}
      \item Prompt injection: attempts to override agent instructions
      \item Data exfiltration: requests for unauthorized data access
      \item Regulatory bypass: social engineering to skip compliance
        checks
      \item Role confusion: attempting to invoke actions outside the
        agent's approval authority
    \end{itemize}
\end{enumerate}

\subsubsection{Generation Algorithm}

\begin{algorithm}
\caption{Ontology-to-Scenario Generation}
\label{alg:scenario-gen}
\begin{singlespace}
\fontsize{10}{14}\selectfont
\begin{algorithmic}[1]
\REQUIRE Industry ontology $\onto$, agent envelope $\env_\agent$
\ENSURE Scenario set $\scenario$
\STATE $\scenario \gets \emptyset$
\FORALL{regulation $\gamma \in \onto.\mathcal{D}.G$}
  \STATE $\scenario_R \gets \textsc{GenRegulatoryScenarios}(
    \gamma, \env_\agent.\Pi)$
  \COMMENT{Generate positive and negative test cases per regulation}
  \STATE $\scenario \gets \scenario \cup \scenario_R$
\ENDFOR
\FORALL{metric $m \in \onto.\mathcal{D}.M$}
  \STATE $\scenario_K \gets \textsc{GenKPIScenarios}(
    m, m.\text{healthy\_range}, m.\text{world\_class})$
  \COMMENT{Test metric calculation and interpretation}
  \STATE $\scenario \gets \scenario \cup \scenario_K$
\ENDFOR
\FORALL{action $a \in \env_\agent.\Pi$}
  \STATE $\scenario_A \gets \textsc{GenAdversarialScenarios}(a)$
  \COMMENT{Boundary conditions, injection, bypass attempts}
  \STATE $\scenario \gets \scenario \cup \scenario_A$
\ENDFOR
\STATE $\scenario \gets \textsc{Deduplicate}(\scenario)$
\RETURN $\scenario$
\end{algorithmic}
\end{singlespace}
\end{algorithm}

Each regulatory constraint $\gamma$ generates at minimum two scenarios:
a \emph{positive case} (the agent should apply the regulation correctly)
and a \emph{negative case} (the agent should detect and reject a
regulatory violation). For regulations with numeric thresholds (e.g.,
BSA/AML \$10,000 reporting threshold), additional \emph{boundary cases}
are generated at $\text{threshold} \pm \epsilon$.

\paragraph{Information hierarchy.}
\Cref{alg:scenario-gen} represents the richest information condition
for scenario generation (G4 in our empirical evaluation,
\Cref{sec:empirical}). We can formalize the information available to
alternative generation strategies as a strict hierarchy:

\begin{equation}
  \text{I}_{\text{G1}} \subset \text{I}_{\text{G2}} \subset
  \text{I}_{\text{G3}} \subset \text{I}_{\text{G4}}
\end{equation}

\noindent where:
\begin{itemize}
  \item $\text{I}_{\text{G1}} = \{r, d\}$: agent role $r$ and
    industry domain name $d$ only (baseline LLM knowledge)
  \item $\text{I}_{\text{G2}} = \text{I}_{\text{G1}} \cup \{P, C\}$:
    adds persona set $P$ and scenario category set $C$
    (behavioral coverage matrix)
  \item $\text{I}_{\text{G3}} = \text{I}_{\text{G1}} \cup
    \{\text{chunks}(\onto)\}$: adds unstructured text extracted from
    the ontology (simulating RAG retrieval)
  \item $\text{I}_{\text{G4}} = \text{I}_{\text{G1}} \cup
    \{\onto' \subseteq \onto\}$: adds the full structured ontology
    (with holdout partition $\onto' = (1 - h) \cdot \onto$ for
    anti-circularity, where $h = 0.30$)
\end{itemize}

\noindent The distinction is \emph{structural}: G3 receives the same ontological \emph{content} as G4 in flattened text form, whereas G4 receives the three-layer structure ($\langle \mathcal{R}, \mathcal{D}, \mathcal{I} \rangle$) that enables \Cref{alg:scenario-gen}'s systematic traversal of regulatory constraints, metrics, and permission boundaries. The pilot study asks whether structure matters above and beyond content.

\subsubsection{Coverage Analysis}

Ontology-grounded generation enables formal coverage analysis:

\begin{Definition}[Regulatory Coverage]
  The regulatory coverage of a scenario set $\scenario$ against
  ontology $\onto$ is:
  \begin{equation}
    \text{RC}(\scenario, \onto) = \frac{|\{\gamma \in \onto.\mathcal{D}.G
      \mid \exists s \in \scenario: s \text{ tests } \gamma\}|}{
      |\onto.\mathcal{D}.G|}
  \end{equation}
\end{Definition}

We validate the coverage metric empirically in \Cref{sec:empirical} against a ground-truth checklist of 125 regulatory requirements (25 per industry) curated from primary sources, comparing G4 to three alternative generation strategies.

\subsection{Trust Certificate}
\label{sec:safety-certificate}

A Trust Certificate is a machine-verifiable attestation of agent safety that binds a specific agent version to its verification evidence.

\subsubsection{Certificate Structure}

\begin{Definition}[Trust Certificate]
  A Trust Certificate $\cert$ for agent $\agent$ is a tuple:
  \begin{equation}
    \cert_\agent = \langle \env_\agent, \scenario, \mathbf{R},
      \verdict, t, \text{sig} \rangle
  \end{equation}
  where:
  \begin{itemize}
    \item $\env_\agent$: The operational envelope under which
      certification was performed
    \item $\scenario$: The scenario set executed during verification
    \item $\mathbf{R}$: The results matrix (per-scenario pass/fail
      with judge evaluations)
    \item $\verdict$: The certification verdict
    \item $t$: The certification timestamp
    \item $\text{sig}$: A cryptographic signature binding the
      certificate to the specific agent version
  \end{itemize}
\end{Definition}

\subsubsection{Verdict Framework}

The certification verdict $\verdict$ is determined by aggregate
simulation results:

\begin{equation}
  \verdict(\mathbf{R}) = \begin{cases}
    \textsc{Approved} & \text{if } \text{pass\_rate}(\mathbf{R})
      \geq \theta_{\text{high}} \\
    \textsc{Conditional} & \text{if } \theta_{\text{low}} \leq
      \text{pass\_rate}(\mathbf{R}) < \theta_{\text{high}} \\
    \textsc{Rejected} & \text{if } \text{pass\_rate}(\mathbf{R})
      < \theta_{\text{low}}
  \end{cases}
\end{equation}

In the \faos{} implementation, $\theta_{\text{high}} = 0.95$ and
$\theta_{\text{low}} = 0.80$. The intermediate verdict
\textsc{Conditional} requires manual review by an L3-authorized
human operator before production deployment.

\subsubsection{Verdict Enforcement: The Simulation Gate}

The simulation gate is an \emph{architectural enforcement point}
that blocks agent deployment based on certification verdict:

\begin{equation}
  \text{deploy}(\agent) = \begin{cases}
    \text{allow} & \text{if } \verdict = \textsc{Approved} \\
    \text{require\_L3\_approval} & \text{if } \verdict =
      \textsc{Conditional} \\
    \text{block} & \text{if } \verdict = \textsc{Rejected}
  \end{cases}
\end{equation}

The gate is designed to operate at the infrastructure level (Rust-native AgentOS runtime) rather than the application layer, preventing bypass by application code, and is environment-aware: enforced in \texttt{production}/\texttt{customer-vpc}, skipped in \texttt{staging}/\texttt{development}.

\subsubsection{Certificate Properties}

A well-formed Trust Certificate must satisfy four framework-level design requirements. These are axiomatic preconditions for certification, not theorems derived from the definition; the pilot study does not empirically validate them ($\theta_{\text{Completeness}}{=}0.95$ is proposed, not achieved by any agent in the pilot).

\begin{Property}[Completeness --- Design Requirement]
  $\text{RC}(\scenario, \onto) \geq 0.95$ for all applicable regulatory frameworks.
\end{Property}

\begin{Property}[Freshness --- Design Requirement]
  $t_{\text{current}} - t \leq \Delta t_{\text{max}}$; expiry forces re-verification on ontology or agent-code change.
\end{Property}

\begin{Property}[Version Binding --- Design Requirement]
  $\text{sig}$ binds $\cert$ to a specific triple---code hash, model version, and ontology version---so any change invalidates $\cert$.
\end{Property}

\begin{Property}[Non-Transferability --- Design Requirement]
  $\cert_{\agent_i}$ is valid only for $\agent_i$; $\agent_j \neq \agent_i \implies \cert_{\agent_i}$ inapplicable.
\end{Property}

\subsection{Implementation Architecture}
\label{sec:simulation-architecture}

An implementation architecture that executes the framework of \Cref{sec:proposed-framework} addresses Gap~(4) of \Cref{sec:gaps}: an infrastructure-level deployment gate that cannot be bypassed by application code. We propose a bilingual design for the \faos{} platform---a Rust-native simulation runner (for execution performance and sandboxing) paired with a Python LLM-as-judge evaluator (for semantic assessment), orchestrated via Temporal workflows for durability. The LLM-as-judge component has been empirically validated in our RA-3 pilot \cite{luong2026neurosymbolic}; the simulation runner and deployment gate are proposed designs.

\subsubsection{Simulation Runner}

The proposed Rust-native runner provides four execution guarantees.
\emph{Concurrency control:} bounded parallelism via
\texttt{tokio::Semaphore(10)} limits concurrent sandbox executions and
prevents resource exhaustion. \emph{Sandbox isolation:} each scenario
executes in an isolated sandbox environment with controlled tool
access. \emph{Timeout enforcement:} per-scenario timeout (default
60\,s, configurable per industry) with deterministic cleanup ensures
no scenario stalls the suite. \emph{Crash recovery:} Temporal workflow
durability ensures simulation state survives runner crashes and
resumes from the last checkpoint.

\subsubsection{LLM-as-Judge Evaluator}

Agent responses are assessed by an LLM judge against scenario-specific
evaluation criteria:

\begin{equation}
  \text{judge}(a, s) = \langle \text{pass} \in \{0, 1\},
    \text{score} \in [0, 1], \text{rationale} \in \text{String} \rangle
\end{equation}

The judge evaluates each response along four dimensions, each phrased
as a verifiable assertion. \emph{Regulatory compliance:} the response
correctly applies all applicable regulations to the scenario.
\emph{Domain accuracy:} domain terms and metrics are used correctly
according to the industry ontology. \emph{Safety boundary:} the
response remains within the agent's operational envelope.
\emph{Adversarial resilience:} the agent resists manipulation attempts
and refuses unauthorized actions.

In our pilot study (\Cref{sec:empirical}), we implement five
specialized judge variants that decompose this general
evaluator into measurable dimensions (\Cref{tab:judge-variants}):

\begin{table}[htbp]
\centering
\caption{LLM-as-Judge variants used in the pilot study.}
\label{tab:judge-variants}
\begin{tabular}{@{}llll@{}}
\toprule
\textbf{Judge} & \textbf{Input} & \textbf{Output} & \textbf{DV} \\
\midrule
RC judge & regulation $+$ scenario & binary $+$ rationale & RC \\
ISS judge & scenario $+$ industry & 1--5 score & ISS \\
AC judge & scenario & category or null & AC \\
FDR-design & fault spec $+$ scenario & binary & FDR-d \\
FDR-exec & fault $+$ agent response & binary & FDR-e \\
\bottomrule
\end{tabular}
\end{table}

\noindent All judge variants use temperature $T=0.0$ for
deterministic evaluation. Each receives a structured rubric
constraining its assessment scope, following the recommendations
of \cite{zheng2023judging} for reducing LLM-as-judge bias.

\subsubsection{Coverage Reporting and Observability}

The coverage reporter produces a per-vertical analysis
\begin{equation*}
  \text{coverage}(\agent, \onto) = \langle \text{RC}_{\text{reg}},\ \text{RC}_{\text{kpi}},\ \text{RC}_{\text{adv}},\ \text{RC}_{\text{total}} \rangle,
\end{equation*}
where each component measures the percentage of ontological elements covered by passing test scenarios; this report is embedded in the Trust Certificate. For operational monitoring the simulation engine exports four Prometheus metrics (total scenarios, failed scenarios, sandbox crashes, per-scenario duration) labeled by verdict, industry, and failure category.

\section{Proposed Evaluation Framework}
\label{sec:eval-framework}

The framework of \Cref{sec:proposed-framework} specifies \emph{what} to verify; the architecture of \Cref{sec:simulation-architecture} specifies \emph{how}. Two complementary evaluation frameworks establish trust in the resulting Trust Certificate. The Quantitative Evaluation Framework (\Cref{sec:formal-extensions}) extends simulation toward model checking and runtime verification to obtain mathematical guarantees; the Empirical Evaluation Framework (\Cref{sec:empirical}) provides statistical evidence through a controlled study and three-model cross-validation. Together they address gaps~(5) and~(6) of \Cref{sec:gaps}.

\subsection{Quantitative Evaluation Framework}
\label{sec:formal-extensions}

Simulation provides statistical confidence, not mathematical guarantees. Three extensions move toward formal agent verification.

\subsubsection{Property Specification Language}

We extend Linear Temporal Logic (LTL) \cite{pnueli1977temporal} with an ontological predicate $\texttt{onto}(d, c)$ asserting concept $c$ in domain $d$:
\begin{equation}
  \prop ::= p \mid \neg\prop \mid \prop \wedge \prop \mid
    \bigcirc\prop \mid \prop \mathcal{U} \prop \mid
    \Box\prop \mid \Diamond\prop \mid
    \texttt{onto}(d, c)
\end{equation}
Given ontology $\onto = \langle \mathcal{R}, \mathcal{D}, \mathcal{I} \rangle$ and agent state $s$ with action--observation pair $(a, o)$:
\begin{equation}
  s \models \texttt{onto}(d, c) \iff
    \exists e \in \text{extract}(s) : e \in
    \text{instances}(\onto.\mathcal{D}_d, c)
\end{equation}
where $\text{extract}(s)$ maps state to domain entities (NER or structured-output parsing) and $\text{instances}(\onto.\mathcal{D}_d, c)$ returns all instances of $c$ in $d$. Instance checking is polynomial-time and decidable under OWL~2~EL/QL; full OWL~2~DL is \textsc{ExpTime}. The FAOS ontologies use a restricted vocabulary amenable to efficient checking.

\begin{tcolorbox}[title=Example Safety Property (BSA/AML),
  colback=red!5, colframe=red!40]
\small
\[
  \prop_{\text{CTR}} : \Box(\texttt{amount} > 10000 \implies \Diamond\, \texttt{file\_ctr})
\]
\emph{``Always: if the transaction amount exceeds \$10K, eventually file a Currency Transaction Report.''} Analogous HIPAA and Fair-Lending properties follow the same pattern with $\texttt{onto}(d, c)$ predicates parameterised by domain.
\end{tcolorbox}

\subsubsection{Bounded Model Checking}

For finite-horizon agents (e.g., single-turn analysis), bounded model checking (BMC) \cite{biere2003bounded} verifies $\text{BMC}(\agent, \prop, k) \iff \forall \tau \in \text{Traces}_k(\agent): \tau \models \prop$ where $\text{Traces}_k$ is all execution traces of length $\leq k$. Because LLM agents are stochastic, we define \emph{probabilistic BMC} using tools such as PRISM \cite{kwiatkowska2011prism}:
\begin{equation}
  \text{PBMC}(\agent, \prop, k, \delta) \iff
    P[\tau \models \prop \mid \tau \in \text{Traces}_k(\agent)]
    \geq 1 - \delta
\end{equation}
yielding a probabilistic guarantee that $\prop$ holds with confidence $1-\delta$ across all traces of bounded length $k$.

\subsubsection{Runtime Verification}

For deployed agents, a runtime monitor \cite{bartocci2018specification} checks safety properties against the live execution trace (\Cref{alg:runtime-monitor}):

\begin{algorithm}
\caption{Runtime Safety Monitor}
\label{alg:runtime-monitor}
\begin{singlespace}
\fontsize{10}{14}\selectfont
\begin{algorithmic}[1]
\REQUIRE Agent $\agent$, safety properties $\Phi$,
  operational envelope $\env_\agent$
\LOOP
  \STATE $e \gets \textsc{ObserveEvent}(\agent)$
  \STATE $\tau \gets \tau \cdot e$
    \COMMENT{Append to trace}
  \FORALL{$\prop \in \Phi$}
    \STATE $v \gets \textsc{Evaluate}(\prop, \tau)$
    \IF{$v = \textsc{Violated}$}
      \STATE $\textsc{Intervene}(\agent, \prop, e)$
        \COMMENT{Pause, rollback, or escalate}
      \STATE $\textsc{EmitEvent}(\text{safety\_violation},
        \prop, e)$
    \ELSIF{$v = \textsc{Inconclusive}$}
      \STATE $\textsc{IncrementWatchdog}(\prop)$
    \ENDIF
  \ENDFOR
\ENDLOOP
\end{algorithmic}
\end{singlespace}
\end{algorithm}

The runtime monitor operates over the existing immutable event
store in \faos{}, which provides append-only, tamper-evident
recording of all agent actions with provenance chains. This
architecture ensures that the monitor has a complete, trustworthy
record of agent behavior.

\subsubsection{The Verification Spectrum}
\label{sec:verification-spectrum}

The three formal techniques above---temporal-logic specification,
bounded model checking, and runtime verification---are not
substitutes but complementary points on a verification spectrum that
trades increasing cost for increasing rigor.
\Cref{tab:verification-spectrum} organizes this spectrum into five
levels, each carrying a specific guarantee and a direct analogy to
established practice in safety-critical software engineering.

\begin{table}[htbp]
\centering
\caption{Agent Verification Spectrum: five levels of pre-deployment
  rigor for enterprise AI agents.}
\label{tab:verification-spectrum}
\begin{tabular}{@{}clp{4cm}l@{}}
\toprule
\textbf{Level} & \textbf{Method} & \textbf{Guarantee} &
  \textbf{Engineering Analogy} \\
\midrule
V1 & Simulation             & Statistical coverage          & Software testing \\
V2 & Probabilistic BMC      & Probabilistic bounds          & Statistical model checking \\
V3 & Runtime monitor        & Live invariant enforcement    & Runtime verification \\
V4 & Bounded model checking & Mathematical proof (bounded)  & Bounded model checking \\
V5 & Theorem proving        & Mathematical proof (unbounded)& Formal verification \\
\bottomrule
\end{tabular}
\end{table}

Target-level selection is governed by regulatory regime, autonomy level (via Property~\ref{prop:autonomy-monotonicity}), and cost-of-failure profile. The framework supports incremental adoption. Organisations may begin at V1 and escalate selected safety-critical properties to V2--V3 without re-architecting the pipeline.

\subsection{Empirical Evaluation Framework}
\label{sec:empirical}

A controlled pilot study tests whether ontology-grounded scenario generation produces test suites measurably superior to alternatives. Conditions span four generation strategies and four regulated industries (Fintech, Banking, Insurance, Healthcare) instantiated as five industry-by-regulatory-regime cells across the United States and Vietnam; the statistical protocol follows the procedure of the companion materials \cite{luong2026neurosymbolic}. We chose this US/Vietnam cross-jurisdictional mix deliberately. The US verticals establish a regulatory baseline familiar to international reviewers (BSA/AML, HIPAA, NAIC), while the Vietnamese verticals stress-test the framework against an underrepresented regulatory regime where (a) Vietnamese-language statutes are not present in most LLM pretraining data, and (b) the regulations were enacted or revised recently (post-2025), post-dating most pretraining cutoffs. This combination tests whether ontology grounding helps even when the LLM lacks prior parametric knowledge of the regulatory regime.

\subsubsection{Enterprise AI Assurance Framework}

Does ontology-grounded scenario generation produce test suites with higher regulatory coverage, fault detection, and industry specificity than baseline, persona-based, or RAG-augmented alternatives? This study tests that question. The independent variable is a within-subjects scenario-generation \emph{condition} with four levels. \textbf{G1 (Baseline)}: the LLM receives only the agent role and industry name. \textbf{G2 (Persona-Scenario Matrix)}: the LLM crosses five personas with six scenario categories (cf.~\asim{} \cite{lyzr2026}). \textbf{G3 (RAG-Augmented)}: the LLM receives eight unstructured text chunks retrieved from the industry ontology, simulating a RAG pipeline. \textbf{G4 (Ontology-Grounded)}: the LLM receives the full three-layer structured ontology following \Cref{alg:scenario-gen}, with 30\% of regulatory constraints held out from the prompt to control circularity (E1). The blocking factor is the industry-by-regulatory-regime cell ($k=5$): US fintech (BSA/AML, KYC), US insurance (NAIC, state DOI), US healthcare (HIPAA, EMTALA), Vietnamese banking (SBV circulars, Decree~116), and Vietnamese insurance (Insurance Business Law 2022, Circular~132). These cells map four regulated industries (Fintech, Banking, Insurance, Healthcare) across two regulatory regimes, yielding cross-regulatory-regime validation. Each condition~$\times$~industry cell is replicated three times at generator temperature $T=0.3$ (E2 protocol fix), producing $4 \times 5 \times 3 = 60$ independently generated test suites of 30 scenarios each (1{,}800 total scenarios).

\subsubsection{Dependent Variables}

Four dependent variables map to the quality dimensions of \Cref{sec:scenario-generation} (regulatory source details for RC are consolidated in \Cref{sec:empirical}'s Ground Truth subsection):

\begin{enumerate}
  \item \textbf{Regulatory Coverage (RC)}: proportion of a 125-item checklist (25 per industry) curated from primary regulatory sources covered by at least one scenario; per-requirement binary, $n{=}125$ per condition.

  \item \textbf{Fault Detection Rate (FDR)}: two-stage assessment (E4)---\emph{FDR-design} asks whether the suite contains a scenario that \emph{should} detect each injected fault; \emph{FDR-execution} asks whether that scenario \emph{does} detect the fault when run against a fault-injected agent. 25 faults across five categories (threshold errors, missing regulations, role-boundary violations, adversarial vulnerabilities, metric-calculation errors); per-fault binary, $n{=}25$.

  \item \textbf{Industry Specificity Score (ISS)}: LLM-as-judge rating on a 1--5 Likert scale per scenario (1 generic, 5 deeply specialized); per-scenario, $n{=}150$.

  \item \textbf{Adversarial Coverage (AC)}: proportion of a six-category adversarial taxonomy covered (prompt injection, data exfiltration, regulatory bypass, role confusion, boundary manipulation, social engineering); per-category, $n{=}6$.
\end{enumerate}

\subsubsection{Evaluation Method}

All assessments use an LLM-as-judge approach (Claude Sonnet~4 at
$T=0.0$ for deterministic evaluation). Each judge call receives a
structured rubric specific to the DV:
\begin{itemize}
  \item \emph{RC judge}: Receives one regulation + one scenario; returns
    binary \texttt{covered/not-covered} with rationale.
  \item \emph{ISS judge}: Receives one scenario + industry; returns
    1--5 score with rationale.
  \item \emph{AC judge}: Receives one scenario; classifies into
    adversarial category or \texttt{null}.
  \item \emph{FDR-design judge}: Receives one fault spec + one scenario;
    returns binary \texttt{would\_detect}.
  \item \emph{FDR-execution judge}: Receives a faulty-agent response +
    expected behavior; returns binary \texttt{fault\_detected}.
\end{itemize}

\subsubsection{Ground Truth and Anti-Circularity Controls}

Regulatory checklists (125 items, 25 per industry) are curated from primary statutory sources---not the FAOS ontology---to avoid circularity: 31~CFR~\S1010--1020 plus OCC 2013-29 and Dodd-Frank \S1071 (fintech); NAIC Models \#785/\#668/\#900, RBC instructions, state DOI filings (insurance); 45~CFR~\S164, 42~USC~\S1395dd, 42~CFR~\S411 (healthcare); SBV Circulars 11/2021, 22/2023, 39/2016 and Decree~116/2013 (Vietnamese banking); Insurance Business Law 2022, Circular~132/2023, Decree~67/2023, AML Law 2022 (Vietnamese insurance). The fault-injection corpus comprises 25 faults (5 per industry) across five categories: threshold errors, missing regulations, role-boundary violations, adversarial vulnerabilities, and metric-calculation errors. Each fault has an injected deviation, ground truth, and expected detection mechanism. Faults are injected into the agent's system prompt, not the scenarios. Vietnamese faults include NPL thresholds (SBV Group~1 $\leq$30 vs.\ correct $\leq$10 days) and solvency-margin confusion (80\% vs.\ 100\%). For E1 anti-circularity, 30\% of G4's regulatory constraints are held out from the generation prompt; the disaggregated seen-vs-unseen RC split will be released with the v0.2 artifact at paper acceptance \cite{faos-research-repo}.

\subsubsection{Statistical Analysis and Hypotheses}

The primary test is Friedman's rank-sum test (nonparametric repeated-measures ANOVA) for each DV, with conditions as treatments and industry-replication combinations as blocks ($\alpha = 0.05$). Post-hoc comparisons use Wilcoxon signed-rank tests with Bonferroni correction for G4 versus each alternative ($m=3$ comparisons); effect size is Kendall's~$W$. \emph{A priori} practical-significance thresholds (E5) are RC~$\geq 15$ pp, FDR~$\geq 2$ additional faults (of 15), ISS~$\geq 1.0$ on the 5-point scale, with AC tested via TOST equivalence at $\pm 15$ pp. Four hypotheses are pre-registered: \textbf{H1}---G4 achieves significantly higher RC than G1/G2/G3; \textbf{H2}---G4 achieves significantly higher FDR; \textbf{H3}---G4 achieves significantly higher ISS; \textbf{H4}---AC is equivalent across conditions (ontology does not uniquely improve adversarial coverage).

\subsubsection{Results}
\label{sec:results}


\begin{table}[htbp]
\centering
\caption{Regulatory Coverage (RC) by condition and industry.
  Values are proportion of 25-item checklist covered ($\pm$ SD across
  3 replications). Bold = highest per industry.}
\label{tab:rc-results}
\begin{tabular}{@{}lcccc@{}}
\toprule
\textbf{Industry} & \textbf{G1 (Base)} & \textbf{G2 (Persona)} &
  \textbf{G3 (RAG)} & \textbf{G4 (Onto)} \\
\midrule
Fintech       & .40$\pm$.00 & .32$\pm$.00 & .37$\pm$.05 & \textbf{.39}$\pm$.06 \\
Insurance     & .39$\pm$.02 & .35$\pm$.06 & .41$\pm$.02 & \textbf{.39}$\pm$.08 \\
Healthcare    & .31$\pm$.02 & .33$\pm$.02 & .36$\pm$.04 & \textbf{.59}$\pm$.14 \\
Banking (VN)  & .53$\pm$.05 & .37$\pm$.06 & \textbf{.57}$\pm$.02 & .49$\pm$.02 \\
Insurance (VN)& .40$\pm$.04 & .28$\pm$.21 & .51$\pm$.08 & \textbf{.56}$\pm$.04 \\
\midrule
\textbf{Mean} & .41$\pm$.08 & .33$\pm$.03 & .45$\pm$.09 & \textbf{.48}$\pm$.09 \\
\bottomrule
\end{tabular}
\end{table}

\begin{table}[htbp]
\centering
\caption{Fault Detection Rate by condition (25 faults across 5 industries).
  FDR-design = suite contains a relevant scenario; FDR-execution = fault
  actually detected at runtime. Mean $\pm$ SD across 15 suites per condition.}
\label{tab:fdr-results}
\begin{tabular}{@{}lcccc@{}}
\toprule
\textbf{Metric} & \textbf{G1} & \textbf{G2} & \textbf{G3} & \textbf{G4} \\
\midrule
FDR-design      & .59$\pm$.09 & .57$\pm$.20 & \textbf{.64}$\pm$.22 & .55$\pm$.14 \\
FDR-execution   & .48$\pm$.18 & .53$\pm$.12 & \textbf{.57}$\pm$.10 & .40$\pm$.11 \\
Gap (d $-$ e)   & +.11 & +.04 & +.07 & +.15 \\
\bottomrule
\end{tabular}
\end{table}

\begin{figure}[htbp]
\centering
\includegraphics[width=0.6\columnwidth]{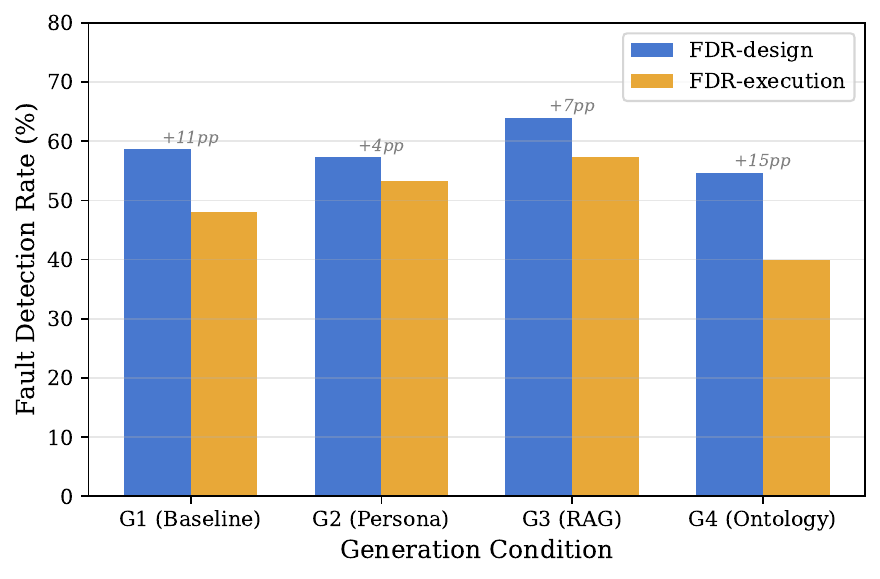}
\caption{Fault Detection Rate: design-time coverage vs.\ runtime
  execution. On the Claude pilot, G4 (Ontology) exhibits the largest gap
  (+15pp), suggesting an apparent \emph{coverage-precision tradeoff} on
  this model; \Cref{sec:crossmodel} shows the gap does not replicate on
  Qwen ($-$16pp) or Gemma (+3pp), so the tradeoff is reported as a
  model-dependent observation rather than an established finding.}
\label{fig:fdr-comparison}
\end{figure}

\begin{table}[htbp]
\centering
\caption{Industry Specificity Score (ISS) by condition and industry.
  Mean score on 1--5 Likert scale ($\pm$ SD across 30 scenarios $\times$
  3 replications). Bold = highest per industry.}
\label{tab:iss-results}
\begin{tabular}{@{}lcccc@{}}
\toprule
\textbf{Industry} & \textbf{G1} & \textbf{G2} & \textbf{G3} & \textbf{G4} \\
\midrule
Fintech       & 4.71$\pm$.08 & 4.57$\pm$.03 & 4.77$\pm$.06 & \textbf{4.79}$\pm$.08 \\
Insurance     & 4.34$\pm$.05 & 4.31$\pm$.07 & 4.20$\pm$.03 & \textbf{4.83}$\pm$.03 \\
Healthcare    & 4.54$\pm$.10 & 4.42$\pm$.12 & 4.56$\pm$.20 & \textbf{4.76}$\pm$.07 \\
Banking (VN)  & 4.72$\pm$.17 & 4.73$\pm$.15 & \textbf{4.84}$\pm$.08 & 4.74$\pm$.13 \\
Insurance (VN)& 4.61$\pm$.07 & 4.63$\pm$.03 & 4.73$\pm$.18 & \textbf{4.74}$\pm$.14 \\
\midrule
\textbf{Mean} & 4.59$\pm$.15 & 4.53$\pm$.17 & 4.62$\pm$.25 & \textbf{4.77}$\pm$.04 \\
\bottomrule
\end{tabular}
\end{table}

\begin{table}[htbp]
\centering
\caption{Statistical significance tests (Friedman rank-sum, $k=4$ conditions). Post-hoc: Wilcoxon signed-rank with Bonferroni correction for G4 vs.\ each alternative. Correction is per-DV ($m{=}3$); a family-wise correction over all primary tests would use $m{\approx}12$ (4~DVs~$\times$~3~post-hocs), which leaves G4~$>$~G2 on RC ($p_c{=}.0006$, significant at $\alpha/12$) and G4~$>$~all on ISS ($p_c{<}.001$, significant at $\alpha/12$) but does not change the non-significance of the other G4 comparisons. Kendall's~$W$ effect sizes remain small ($W < 0.07$) across all DVs.}
\label{tab:stats}
\begin{tabular}{@{}lcccl@{}}
\toprule
\textbf{DV} & \textbf{$\chi^2$} & \textbf{$p$-value} &
  \textbf{Kendall's $W$} & \textbf{Verdict (per-DV Bonferroni)} \\
\midrule
RC ($n=125$)  & 15.40 & \textbf{.0015}$^{**}$ & .041 &
  G4 $>$ G2 ($p_c{=}.0006$)$^{***}$; G4 vs.\ G1, G3 n.s. \\
FDR-d ($n=25$) & 0.63 & .891 & .008 & n.s. \\
ISS ($n=150$) & 29.45 & \textbf{$2{\times}10^{-6}$}$^{***}$ & .065 &
  G4 $>$ all ($p_c{<}.001$)$^{***}$ \\
AC ($n=6$)    & 0.07 & .995 & .004 & Equivalent (as predicted) \\
\bottomrule
\end{tabular}
\end{table}

The results (\Cref{tab:stats}) support the industry-specificity hypothesis (H3) and the adversarial-equivalence hypothesis (H4), partially support the regulatory-coverage hypothesis (H1, robust only against the persona baseline G2), do not support the fault-detection hypothesis (H2), and surface an unexpected fault-detection pattern.

\emph{H1 (RC): Supported against G2.} Ontology-grounded generation (G4)
leads with mean regulatory coverage of 48.3\%, well above persona-based
generation (G2, 33.1\%; post-hoc $p_c = 0.0006$).
The largest per-industry point-estimate gains over G1 occur in healthcare
(+28pp) and Vietnamese insurance (+16pp), suggesting that ontology value
scales with regulatory complexity; these are descriptive contrasts only,
as the G4 vs.\ G1 omnibus comparison is not Bonferroni-significant
($p_c = 0.243$). The G4 vs.\ G3 difference (+3.7pp) is likewise not
statistically significant after correction ($p_c = 1.0$), indicating that
\emph{structured} ontology and \emph{unstructured} RAG chunks produce
similar coverage when the underlying content is equivalent. We discuss
this finding further in \Cref{sec:discussion}. A surprising operational finding emerged in the Vietnamese banking cell, where G3 (RAG) achieved a higher mean regulatory coverage than G4 (0.57 vs. 0.49). The published coverage summaries show that, across the three Vietnamese-banking replications, G3 covered 15 unique checklist requirements while G4 covered 14; G3 uniquely covered requirements on customer identity verification, related-party credit exposure, medium- and long-term lending limits, and credit assessment, while G4 uniquely covered cash-transaction reporting, structuring detection, and SBV reporting. This exposes a more cautious trade-off: structured ontology context can improve domain specificity while still narrowing coverage breadth in regulatory regimes whose obligations are distributed across many circulars.

\emph{H2 (FDR): Not supported.} Fault detection rates (\Cref{tab:fdr-results}) show no
significant differences across conditions ($p = .891$). All conditions
generate scenarios capable of detecting approximately 55--64\% of
injected faults at the design level. However, on the Claude pilot G4
shows the lowest \emph{execution} FDR (40\%) and largest design-execution
gap (+15pp; \Cref{fig:fdr-comparison}), suggesting that ontology-generated scenarios, while
regulatory-comprehensive, may be less effective at \emph{triggering}
specific fault behaviors at runtime. On Claude this pattern resembles a
\textbf{coverage-precision tradeoff}: breadth (regulatory coverage)
purchased at the expense of depth (fault-triggering precision). However,
\Cref{sec:crossmodel} shows the gap does not replicate on Qwen
($-$16pp) or Gemma (+3pp), so we report the tradeoff as a
model-dependent, exploratory observation rather than an established finding.

\emph{H3 (ISS): Strongly supported.} G4 produces the most
industry-specific scenarios (4.77/5.0), well above all
alternatives: G1 ($p_c = 3.2 \times 10^{-5}$), G2 ($p_c < 10^{-5}$),
and G3 ($p_c = 0.0008$). The G4-over-G3 ISS gap is significant despite similar RC scores, indicating that structured ontology injection produces scenarios that are regulatory-relevant and domain-specific in formulation.

\emph{H4 (AC): Supported.} Adversarial coverage is equivalent across
conditions ($p = .995$, $W = .004$), confirming that adversarial
scenario generation depends on explicit prompting rather than domain
knowledge source. All conditions achieve 88--91\% coverage of the
six-category adversarial taxonomy.

A specific instance of these two patterns surfaced in the Vietnamese
insurance cell. A risk-classification and capital-adequacy agent
operating under Circular~06/2024/TT-BTC \cite{vncircular062024}
showed the strongest ontology advantage in the Vietnam-regime subset:
G4 reached the highest RC among the four conditions (0.56), ahead of
G3~RAG (0.51), G1 (0.40), and G2 (0.28).
This expansion drove the +16~pp RC gain over G1 (G4~RC~$=0.56$
vs.\ G1~RC~$=0.40$) reported in \Cref{tab:rc-results}. Yet on
industry specificity, G4~(4.74) and G3~RAG~(4.73) finished essentially
tied in this cell---well above persona-based G2, but with the
ontology's marginal advantage over chunked-RAG largely collapsed
(\Cref{tab:iss-results}). The combined pattern suggests an
interpretation we did not anticipate at pre-registration: ontology
grounding earns its keep on regulatory coverage when the relevant
statutes are diffuse or cross-referenced, but its industry-specificity
advantage saturates once the underlying regulatory text is itself
sufficiently dense---as Circular~06/2024 is. Future work should map
this saturation curve explicitly: ontology grounding likely has the
highest marginal value in jurisdictions or verticals where the
regulatory text density is low and cross-reference complexity is high.

\subsubsection{Cross-Model Validation}
\label{sec:crossmodel}

A central threat to the preceding results is that the ontology advantage may reflect model-specific parametric knowledge rather than the structural contribution of ontological context. The full 60-suite experiment was replicated with two additional generator models---Qwen~2.5 72B~Instruct (OpenRouter) and Gemma~4 26B (Google AI Studio)---while holding the Claude Sonnet~4 judge fixed at $T{=}0.0$ to preserve measurement consistency. The identical
experimental design (4~conditions $\times$ 5~industries $\times$
3~replications $\times$ 30~scenarios) yields 1,800 scenarios per
model and 5,400 total across all three.

\Cref{tab:crossmodel} summarizes the G4 (ontology) performance
across generator models.

\begin{table}[ht]
\centering
\caption{Cross-model comparison of G4 (ontology-grounded) generation.
Judge is Claude Sonnet~4 ($T{=}0.0$) for all models.}
\label{tab:crossmodel}
\small
\begin{tabular}{lcccccc}
\toprule
& \multicolumn{2}{c}{\textbf{RC}} & \multicolumn{2}{c}{\textbf{ISS}} & \multicolumn{2}{c}{\textbf{FDR}} \\
\cmidrule(lr){2-3} \cmidrule(lr){4-5} \cmidrule(lr){6-7}
\textbf{Generator} & G4 & $\Delta$(G4$-$G2) & G4 & $\chi^2$ & Design & Exec \\
\midrule
Claude Sonnet~4 & 48.3\% & +15.2pp & 4.77 & 29.4*** & 54.7\% & 40.0\% \\
Gemma~4 26B     & 40.3\% & +11.2pp & 4.68 & 70.1*** & 53.3\% & 50.7\% \\
Qwen~2.5 72B    & 30.1\% & +14.4pp & 4.37 & 103.2*** & 40.0\% & 56.0\% \\
\bottomrule
\multicolumn{7}{l}{\footnotesize ***\,$p < .001$ (Friedman omnibus for ISS H3)}
\end{tabular}
\end{table}

\begin{figure}[htbp]
\centering
\subfloat[RC by condition across three models. G4 (ontology) leads or ties for all models; the G4--G2 gap is the statistically-robust effect, while G4--G1 and G4--G3 gaps are not Bonferroni-significant.\label{fig:crossmodel-rc}]{%
  \includegraphics[width=0.48\columnwidth]{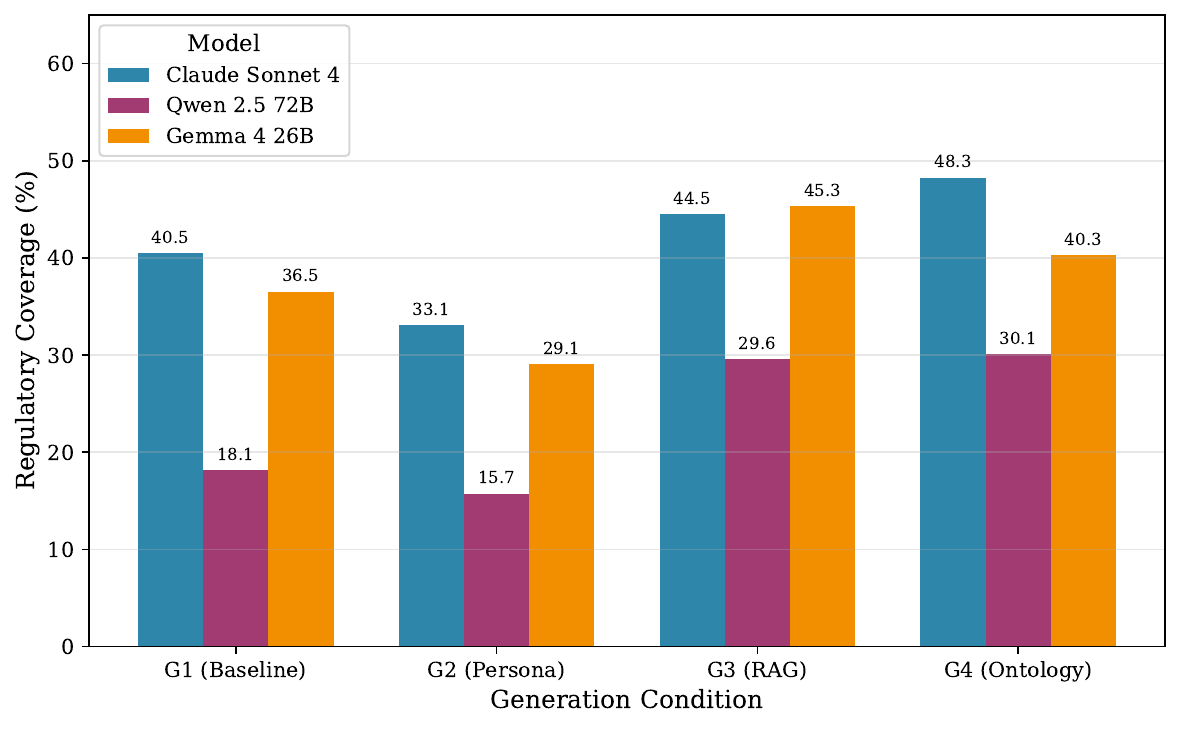}}
\hfill
\subfloat[G4 RC by industry $\times$ model. Claude leads on absolute RC, while ontology uplift is largest for Qwen---consistent with the Inverse Parametric Knowledge Effect.\label{fig:crossmodel-rc-heatmap}]{%
  \includegraphics[width=0.48\columnwidth]{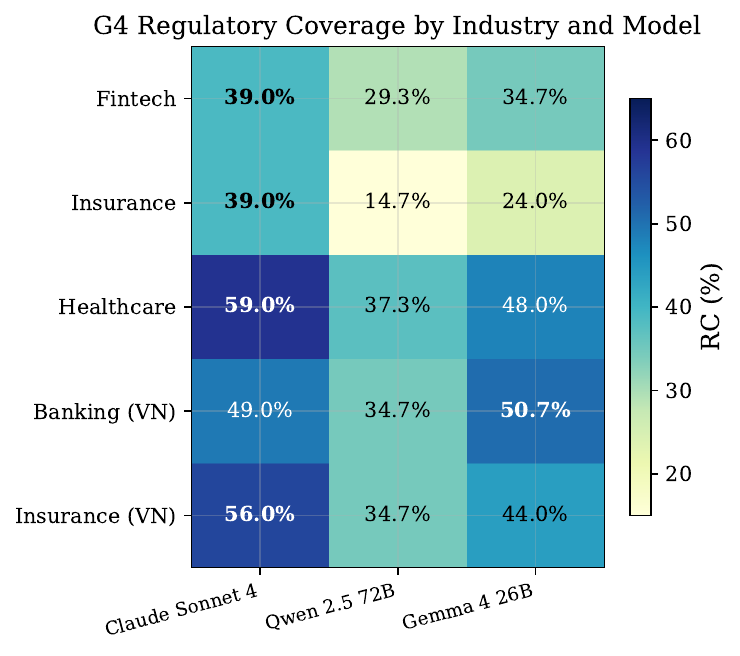}}
\caption{Cross-model replication of ontology advantage: (\textbf{a}) by generation condition, (\textbf{b}) by industry $\times$ generator model.}
\label{fig:crossmodel}
\end{figure}

\paragraph{H5 (RC advantage replicates).}
All three models show G4 significantly exceeding G2 on regulatory
coverage: Claude ($\Delta{=}+15.2$pp, $p_c{=}.0006$), Gemma
($\Delta{=}+11.2$pp, $p_c{=}.009$), and Qwen
($\Delta{=}+14.4$pp, $p_c{=}.005$). The ontology advantage is
\emph{universal}---not an artifact of Claude's parametric knowledge (\Cref{fig:crossmodel-rc}; full cross-model figure: \Cref{fig:crossmodel}).

\paragraph{H6 (ISS advantage replicates).}
G4 achieves the highest industry specificity score across all three
models: Claude (4.77/5.0), Gemma (4.68/5.0), and Qwen (4.37/5.0),
each with omnibus $p < 10^{-5}$. For Qwen, all three post-hoc
comparisons (G4 vs.\ G1, G2, G3) are significant at Bonferroni-corrected
$p < .001$. The ISS gap between models is smallest for G4 (range:
0.40 across models) versus G1 (range: 0.42), suggesting that
\emph{ontology partially compensates for model capability differences}.

\paragraph{H7 (Coverage-precision tradeoff).}
The FDR design--execution gap is model-dependent: Claude +14.7pp (design overestimates execution), Gemma +2.7pp (near-zero), Qwen $-$16.0pp (execution outperforms design). The Claude-only coverage-precision tradeoff is therefore \emph{not universal}, but a model-dependent interaction between ontological structure and the generator's ability to translate regulatory-coverage scenarios into fault-triggering cases.

\paragraph{Exploratory three-model observation.}
Absolute G4 RC orders Claude $>$ Gemma $>$ Qwen (48.3 $>$ 40.3 $>$ 30.1\%); the G4$-$G1 \emph{uplift} reverses: Qwen (+12.0pp) $>$ Claude (+7.7pp) $>$ Gemma (+3.7pp). This is directionally consistent with the Inverse Parametric Knowledge Effect \cite{luong2026neurosymbolic} but rests on $n{=}3$ non-monotone capability points (Gemma 26B vs Qwen 72B) and uses in-study G1 RC as the capability axis, which makes this capability--uplift relationship partially tautological. We therefore flag the gradient as an exploratory observation motivating $\geq$6-model replication, not a confirmed finding (\Cref{fig:crossmodel-rc-heatmap}).

\section{Analysis, Governance Implications, and Limitations}
\label{sec:discussion}

\subsection{Ontology as Specification Language}

Enterprise ontologies serve a \emph{triple role} across the agent lifecycle: \textbf{grounding} (input context that improves agent reasoning, validated in the RA-3 companion \cite{luong2026neurosymbolic}), \textbf{specification} (the test-scenario source via \Cref{alg:scenario-gen}, validated in \Cref{sec:empirical}), and \textbf{oracle} (the evaluation rubric against which the LLM judge assesses responses). This extends the model-based testing paradigm \cite{utting2012taxonomy}---in which the model serves as specification and oracle---by adding the grounding role: a better ontology yields both better agents \emph{and} stronger verification.

\subsection{Comparison with Lyzr A-SIM}

Ontology-grounded generation differs from Lyzr's persona-scenario matrix along several dimensions (\Cref{tab:comparison}):

\begin{table}[htbp]
\centering
\caption{Comparison: Ontology-Grounded vs. Persona-Scenario
  Simulation.\textsuperscript{$\dagger$}}
\label{tab:comparison}
\begin{tabular}{@{}lcc@{}}
\toprule
\textbf{Dimension} & \textbf{Persona-Scenario} & \textbf{Ontology-Grounded} \\
\midrule
Scenario source & Manual + generic & Auto-generated from ontology \\
Industry specificity & Low & High (22 verticals) \\
Regulatory coverage & Incidental & Systematic \\
Coverage metric & Scenario count & Regulatory coverage \% \\
Maintenance & Manual updates & Ontology-grounded updates \\
Formal properties & None & LTL-based specification \\
Certification & Pass/fail & Graduated (App/Cond/Rej) \\
\bottomrule
\end{tabular}

\smallskip
\noindent\textsuperscript{$\dagger$}\footnotesize{Ontology-Grounded column reflects the proposed architecture; the current \faos{} implementation uses template-based generation with manual curation, with full algorithmic generation per \Cref{alg:scenario-gen} planned.}
\end{table}

\noindent The distinction is not architectural but one of \emph{unit of analysis}: persona-scenario matrices optimize for \emph{behavioral coverage} (diverse interaction patterns); ontology-grounded generation optimizes for \emph{regulatory coverage} (verified compliance with the regulatory corpus). The empirical results (\Cref{sec:empirical}) confirm the prediction: G2 achieves comparable AC (88\% vs.\ G4's 91\%, $p{=}.995$) but significantly lower RC (33.1\% vs.\ 48.3\%, $p_c{=}.0006$)---behaviorally diverse yet regulatory-shallow.

\subsection{Governance Framework Alignment}

The 2025--2026 maturation of agent governance provides external validation for the design choices. OWASP's Top~10 for Agentic Applications \cite{owasp2025agentic}---goal hijacking (ASI01), tool misuse (ASI02), rogue agents (ASI10)---maps onto the scenario-generation taxonomy (\Cref{sec:scenario-generation}): regulatory scenarios detect goal deviation, operational scenarios exercise tool boundaries, adversarial scenarios simulate rogue behavior. NIST's Agent Standards Initiative \cite{nist2026agents} calls for identity, security, and monitoring standards; the Trust Certificate (\Cref{sec:safety-certificate}) implements the monitoring pillar and the operational envelope (\Cref{sec:operational-envelope}) formalizes risk management. The Microsoft Agent Governance Toolkit \cite{microsoft2026agt} enforces policy \emph{during} execution; the present framework is its pre-deployment complement---ontology-grounded simulation decides \emph{whether} to deploy, runtime governance constrains \emph{how}---together forming defense-in-depth.

\subsection{Compositional Verification}
\label{sec:compositional}

Enterprise deployments rarely involve single agents: the \faos{} platform supports agent teams via the A2A protocol and LangGraph orchestration, raising the \emph{compositional verification problem}---given individually certified agents $\agent_1,\ldots,\agent_n$, what holds for their composition? Classical assume-guarantee reasoning \cite{clarke2018handbook} provides a foundation (if each $\agent_i$ is verified under environment assumption $A_i$ and $\bigwedge_i \Phi_i \implies \bigwedge_j A_j$, system properties hold), but LLM agents violate its determinism premise: identical messages can yield different outputs. Three directions emerge. First, \emph{probabilistic assume-guarantee} with $P[\Phi_i] \geq 1-\delta_i$ derives system confidence from component confidence through independence or correlation bounds. Second, \emph{envelope intersection} composes the system envelope as $\bigcap_i \env_{\agent_i}$. Third, \emph{protocol-level verification} checks message types, sequencing, and escalation rules with A2A as a checkable contract, independent of agent internals. Full treatment is deferred to future work.

\subsection{Limitations}
\label{sec:limitations}

Three limitations bound the conclusions of this work. \emph{First, ontology completeness:} verification coverage is bounded by ontology coverage---a regulation absent from the ontology creates a systematic blind spot, placing a continuing curation burden on enterprise adopters as a precondition for trust. \emph{Second, LLM-as-judge construct validity:} the evaluation pipeline uses Claude Sonnet~4 as both generator and fixed judge ($T{=}0.0$), introducing potential self-enhancement bias that cross-generator replication alone cannot eliminate (see \Cref{sec:crossmodel}); a related concern is that the same author designed both the FAOS ontology and the regulatory checklist used as ground truth, making the anti-circularity control (E1) a pseudo-control rather than a fully independent check. Deterministic evaluation, ontology-derived binary rubrics, programmatic threshold checks, and three-model cross-validation partially mitigate these risks, but inter-judge and inter-rater triangulation remain the dominant residual concerns. \emph{Third, statistical confidence versus formal guarantees:} simulation provides practical assurance at verification levels V1--V3, but higher levels (V4--V5) may be computationally intractable for agents with large state spaces (\Cref{sec:formal-extensions}). The graduated verdict thresholds ($\theta_{\text{high}}{=}0.95$, $\theta_{\text{low}}{=}0.80$) are illustrative engineering parameters, not yet calibrated against real-world deployment incident rates; no agent in the pilot corpus satisfied $\theta_{\text{high}}$.

These limitations chart a concrete future-work agenda. Inter-rater validation of the 125-item regulatory checklist by non-author regulatory experts on a stratified random sample, paired with non-Anthropic judge triangulation (GPT-4o or Gemini) on the same sub-sample and reported judge--judge agreement metrics for each dependent variable, will quantify residual self-enhancement bias and re-anchor effect-size estimates accordingly. Scaling cross-model replication to six or more LLM families under an independent capability proxy (MMLU score or training-token count) will test whether the cross-model coverage-precision pattern is intrinsic to the ontology approach or an artifact of the present three-model sample; the disaggregated seen-vs-unseen regulatory-coverage split, released with the v0.2 artifact at paper acceptance, supports this scaling exercise. Runtime verification (V3) and probabilistic bounded model checking (V2) on production-deployed agents will move the framework from \emph{proposed} to \emph{delivered} attestation. The framework also extends to multi-agent settings (\Cref{sec:compositional}) through probabilistic assume-guarantee contracts over the A2A protocol, broadening the certification surface to agent teams. Calibrating the verdict thresholds against actual deployment incident rates and publishing a Trust Certificate registry that links specific agent versions to their verification evidence will close the gap between the proposed framework and an operational deployment gate. These directions shift the state of practice from reactive runtime control to systematic pre-deployment assurance for enterprise Agentic AI; \Cref{sec:conclusion} synthesizes the full contribution.

\section{Conclusions}
\label{sec:conclusion}

This study proposes a pre-deployment verification framework for enterprise AI agents, addressing the verification gap framed in \Cref{sec:introduction}. The framework combines three things: ontology-grounded scenario generation, operational-envelope specification, and a proposed Trust Certification model for deployment assurance in regulated environments. The \emph{Agent Operational Envelope} sets the scope an agent is evaluated and authorised to operate within. The \emph{ontology-to-scenario generation pipeline} (\Cref{alg:scenario-gen}) derives regulatory, operational, and adversarial test cases. The \emph{machine-verifiable Trust Certification architecture} has graduated deployment verdicts and a five-level verification spectrum (\Cref{sec:verification-spectrum}) that extends toward formal methods.

The empirical results (\Cref{sec:results}) show that ontology-grounded generation improves regulatory coverage and domain specificity over persona-based simulation. In a controlled pilot spanning four regulated industries (Fintech, Banking, Insurance, Healthcare) instantiated as five industry-by-regulatory-regime cells, 1{,}800 scenarios, 125 primary-source regulatory requirements, and 25 injected faults, the ontology-grounded approach (G4) achieved 48.3\% regulatory coverage versus 33.1\% for the dominant persona-scenario baseline (G2; $p_c{=}.0006$) and the highest industry specificity (4.77/5.0; $p{=}2{\times}10^{-6}$), while the regulatory-coverage advantage over the plain-prompt baseline (G1) and RAG-augmented prompting (G3) is not Bonferroni-robust. Cross-validation across three distinct LLM families (Claude Sonnet~4, Qwen~2.5 72B, and Gemma~4 26B; 5{,}400 total scenarios) replicated the core persona-versus-ontology pattern, alongside an exploratory, model-dependent coverage-precision observation on fault detection, suggesting that the observed gains arise from structural properties of the verification methodology rather than from a single model family.

At the same time, the study deliberately distinguishes statistical assurance from formal guarantees. The proposed Trust Certification framework remains conceptual at higher verification levels, ontology completeness constrains verification coverage, and the fixed LLM-as-judge design leaves residual self-enhancement bias unresolved. Four follow-ups address these limitations: independent evaluator triangulation, broader cross-model replication under an independent capability proxy, probabilistic verification for multi-agent systems through assume-guarantee contracts over the A2A protocol, and progression from simulation-based evidence toward formally grounded deployment assurance.

More broadly, the results suggest that enterprise AI governance may require a transition from reactive runtime control toward systematic pre-deployment verification. Current enterprise practice largely assumes that monitoring, guardrails, and human oversight can compensate for insufficient pre-deployment assurance. This work argues instead for a verification-first paradigm in which autonomous agents are evaluated against explicit operational, regulatory, and safety constraints before deployment. In this context, ontology-grounded simulation represents a practical and scalable step toward enterprise-grade trust infrastructure for Agentic AI systems. We expect \faos{} deployments to support the full governance and conformity-assessment requirements under Vietnam's Law~No.~134/2025/QH15 on Artificial Intelligence \cite{vnailaw2025} once the grace period for financial-services AI closes in September~2027. The framework already produces the artifacts that map directly to that regulatory ask: scenario-coverage records keyed to specific regulatory clauses, fault-detection traces from the simulation gate, and signed Trust Certificates bound to verification evidence---an audit trail that runtime-monitoring-only approaches must improvise post-hoc.

\section*{Author Contributions}
Conceptualization, T.L.T.\ and A.S.; methodology, T.L.T.\ and A.S.; software, T.L.T.; validation, T.L.T.\ and A.S.; formal analysis, T.L.T.; investigation, T.L.T.; resources, T.L.T.; data curation, T.L.T.; writing---original draft preparation, T.L.T.; writing---review and editing, T.L.T.\ and A.S.; visualization, T.L.T.; supervision, A.S.; project administration, T.L.T.

\section*{Funding}
This research received no external funding.

\section*{Data Availability Statement}
The dataset comprises 5,400 generated scenarios across three generator models (Claude Sonnet~4, Qwen~2.5 72B, Gemma~4 26B), each contributing 1,800 scenarios from 60 test suites (4~conditions $\times$ 5~industries $\times$ 3~replications $\times$ 30~scenarios). Per-model assessments include 125 regulatory coverage evaluations per condition, 25 fault injection results, and 150 industry specificity scores per condition---all judged by a fixed Claude Sonnet~4 evaluator ($T{=}0.0$). Analysis scripts, ontology context files, regulatory checklists, fault definitions, aggregated result summaries, and cross-model analysis code are publicly available at the FAOS Research repository \citep{faos-research-repo}. Raw generated-scenario transcripts and judge-level logs are archived at Zenodo \citep{faos-ra6-data} (DOI:~10.5281/zenodo.20484582) and will be released there upon publication.

\section*{Acknowledgments}
The authors thank the FAOS team for platform access and ontology content used in the empirical evaluation.

\section*{Conflicts of Interest}
Author T.L.T.\ is the founder of FAOSX.AI, which develops the FAOS platform whose verification framework is evaluated in this study. This relationship had no role in the study design; in the collection, analysis, or interpretation of data; or in the decision to publish; and all results, including non-significant comparisons, are reported in full. Author A.S.\ declares no conflict of interest.

\bibliography{references}

\end{document}